\documentclass{new_tlp}
\usepackage{mathptmx}

\usepackage{amsmath}
\usepackage{amsbsy}
\usepackage{url}

\newif\ifdraft

\draftfalse

\ifdraft

\else

\newcommand\comment[1]{}

\fi

\usepackage{xspace}
\usepackage{latexsym}
\usepackage[usenames]{xcolor}
\usepackage{float}
\usepackage{longtable}
\usepackage{pdfsync}
\usepackage{txfonts}
\usepackage{enumitem}

\usepackage{framed}
\usepackage{amssymb}
\usepackage{latexsym}
\usepackage{ifthen}

\newcommand\nameFont[1]{{\fontencoding{OT1}\fontfamily{cmss}\selectfont{#1}}}
\def\nf{\nameFont}

\def\langname{\nf{ASP-Core-2}\xspace}
\def\aspcore{\nf{ASP-Core}\xspace}

\newcommand\quo[1]{{``{#1}"}}
\newcommand\nop[1]{}
\newcommand\naf{{\ensuremath{{\mathbf{ not\ }}}}}

\newcommand\AS{\ensuremath{\mathit{AS}}}
\newcommand\Ans{\ensuremath{\mathit{Ans}}}

\newcommand{\aggr}[1]{\ensuremath{\mathrm{\# #1}}}

\newcommand{\Colon}{\ensuremath{\!:\!}}

\newcommand{\p}{\ensuremath{{P}}}
\newcommand{\grnd}{\ensuremath{grnd}\xspace}
\newcommand{\GP}{\ensuremath{\grnd(\p)}\xspace}
\newcommand{\BP}{\ensuremath{B_{\p}}\xspace}
\newcommand{\UP}{\ensuremath{U_{\p}}\xspace}

\newenvironment{indentnew}[1]%
{\begin{list}{}%
         {\setlength{\leftmargin}{#1}}%
         \item[]%
} {\end{list}}

\newcommand\bcmdtab{\noindent\bgroup\tabcolsep=0pt%
  \begin{tabular}{@{}p{10pc}@{}p{20pc}@{}}}
\newcommand\ecmdtab{\end{tabular}\egroup}

  \title[ASP-Core-2 Input Language Format]
        {ASP-Core-2 Input Language Format}

  \author[Calimeri et al.]
         {FRANCESCO CALIMERI\\
          DeMaCS, Universit{\`a} della Calabria, Italy\\
          \email{calimeri@mat.unical.it}
          \and
          WOLFGANG FABER, MARTIN GEBSER\\
          Institut f\"ur Angewandte Informatik,  Alpen-Adria-Universit\"at, Klagenfurt, Austria\\
          \email{\{Wolfgang.Faber,Martin.Gebser\}@aau.at}
          \and
          GIOVAMBATTISTA IANNI\\
          DeMaCS, Universit{\`a} della Calabria, Italy\\
          \email{ianni@mat.unical.it}
          \and
          ROLAND KAMINSKI\\
          Institute of Computer Science, University of Potsdam, Germany\\
          \email{kaminski@cs.uni-potsdam.de}
          \and
          THOMAS KRENNWALLNER\\
          XIMES GmbH, Vienna, Austria\\
          \email{tk@postsubmeta.net}
          \and
          NICOLA LEONE\\
          DeMaCS, Universit{\`a} della Calabria, Italy\\
          \email{leone@mat.unical.it}
          \and
          MARCO MARATEA\\
          DIBRIS, University of Genova, Italy\\
          \email{marco@dibris.unige.it}
          \and
          FRANCESCO RICCA\\
          DeMaCS, Universit{\`a} della Calabria, Italy\\
          \email{ricca@mat.unical.it}
          \and
          TORSTEN SCHAUB\\
          Institute of Computer Science, University of Potsdam, Germany\\
          \email{schaub@cs.uni-potsdam.de}}

\jdate{March 2003}
\pubyear{2003}
\pagerange{\pageref{firstpage}--\pageref{lastpage}}
\doi{S1471068401001193}

\begin{document}

\label{firstpage}

\maketitle

  \begin{abstract}
    Standardization of solver input languages has been a main driver for the growth of several areas within knowledge representation and reasoning, fostering the exploitation in actual applications. In this document we present the \langname standard input language for Answer Set Programming, which has been adopted in ASP Competition events since 2013.
  \end{abstract}



\section{Introduction}

The process of standardizing the input languages of solvers for knowledge representation and reasoning research areas has been of utmost importance for the growth of the related research communities: this has been the case for, e.g., the CNF-DIMACS format for SAT, then extended to describe input formats for Max-SAT and QBF problems, the OPB format for pseudo-Boolean problems, somehow at the intersection between the CNF-DIMACS format and the LP format for Integer Linear Programming, the XCSP3 format for CP solving,  SMT-LIB format for SMT solving, and the STRIPS/PDDL language for automatic planning. The availability of such common input languages have led to the development of efficient solvers in different KR communities, through a series of solver competitions that have pushed the adoption of these standards. The availability of efficient solvers, together with a presence of a common interface language, has helped the exploitation of these methodologies in applications.

The same has happened for Answer Set Programming (ASP) \cite{BrewkaET11}, a well-known approach to knowledge representation and reasoning with roots in the areas of logic programming and non-monotonic reasoning \cite{gelf-lifs-91}, through the development of the \aspcore language \cite{languageformat}.
The first \aspcore version was a rule-based language whose syntax stems from plain Datalog and Prolog, and was a conservative extension to the non-ground case of the {\tt Core} language adopted in the First ASP Competition held in 2002 during the Dagstuhl Seminar ``Nonmonotonic Reasoning, Answer Set Programming and Constraints''\footnote{\url{https://www.dagstuhl.de/en/program/calendar/semhp/?semnr=02381}.}. It featured a restricted set of constructs, i.e. disjunction in the rule heads, both strong and negation-as-failure negation in rule bodies, as well as non-ground rules.

In this document we present the latest evolution of \aspcore, namely \langname, which currently constitutes the standard input language of ASP solvers adopted in the ASP Competition series since 2013 \cite{CalimeriIR14,CalimeriGMR16,GebserMR17,GebserMR17b}.
\langname substantially extends its predecessor by incorporating many language extensions that became mature and widely adopted over the years in the ASP community, such as aggregates, weak constraints, and function symbols.
The ASP competition series pushed its adoption, and significantly contributed both to the availability of efficient solvers for ASP \cite{LierlerMR16,GebserLMPRS18} and to the exploitation of the ASP methodology in academic and in industrial applications \cite{ErdemGL16,leon-ricc-2015-rr,GebserOSR18}.
In the following, we first present syntax and semantics for the basic building blocks of the language, and then introduce more expressive constructs such as choice rules and aggregates, which help with obtaining compact problem formulations. Eventually, we present syntactic restrictions for the use of \langname in practice. 

\section{ASP-Core-2 Language Syntax}
\label{sec:language}

For the sake of readability, the language specification is herein given in the traditional mathematical notation.
A lexical matching table from the following notation to the actual raw input format is provided in Section \ref{sec:lex}.

\paragraph{\bf Terms.}
Terms are either {\em constants}, {\em variables}, {\em arithmetic terms} or {\em functional terms}.
Constants can be either {\em symbolic constants} (strings starting with some lowercase letter),
{\em string constants} (quoted strings) or {\em integers}.
Variables are denoted by strings starting with some uppercase letter.
An {\em arithmetic term} has form $-(t)$ or $(t \diamond u)$
for terms $t$ and $u$ with $\diamond \in \{ \quo{+},\quo{-},\quo{*}, \quo{/} \}$;
parentheses can optionally be omitted in which case standard operator precedences apply.
Given a {\em functor}~$f$ (the {\em function name}) and terms $t_1,\dots,t_n$,
the expression
$f(t_1,\dots,t_n)$ is a {\em functional term} if $n > 0$,
whereas $f()$ is a synonym for the symbolic constant~$f$.

\paragraph{\bf Atoms and Naf-Literals.}
A {\em predicate atom} has form $p(t_1,\dots,t_n)$,
where $p$ is a {\em predicate name}, $t_1, \dots, t_n$ are terms and $n \geq 0$ is the arity of the predicate atom;
a predicate atom $p()$ of arity $0$ is likewise represented by its predicate name~$p$ without parentheses.
Given a predicate atom~$q$,
$q$ and $\neg q$ are {\em classical atoms}.
A {\em built-in atom} has form $t \prec u$
for terms $t$ and $u$ with
${\prec} \in \{ \quo{<}, \quo{\leq}, \quo{=}, \quo{\neq},\allowbreak \quo{>}, \quo{\geq} \}$.
Built-in atoms~$a$ as well as the expressions
$a$ and $\naf a$ for a classical atom~$a$
are {\em naf-literals}. \begin{nop}{%
\footnote{Note that (negative) naf-literals $\naf a$
can only be constructed from classical atoms~$a$.}}\end{nop}
%
%

\paragraph{\bf Aggregate Literals.}
An {\em aggregate element} has form
\[ t_1, \dots, t_m : l_1, \dots, l_n \]
where $t_1, \dots, t_m$ are terms
\comment{MG: In general, $t_1, \dots, t_m$ are terms.
             Any restriction, like the one to variables and/or constants,
             should better go to Section~\ref{sec:aspcompnotes}. GB: done}%
and $l_1, \dots, l_n$ are naf-literals for $m\geq0$ and $n\geq0$.

An {\em aggregate atom} has form
\[ \aggr{aggr}\ E \prec u \]
where
$\aggr{aggr} \in \{ \quo{\aggr{count}}, \quo{\aggr{sum}}, \quo{\aggr{max}},\allowbreak \quo{\aggr{min}} \}$
is an {\em aggregate function name},
${\prec} \in \{ \quo{<},\linebreak[1] \quo{\leq},\linebreak[1] \quo{=},\linebreak[1] \quo{\neq},\linebreak[1] \quo{>},\linebreak[1] \quo{\geq} \}$
is an {\em aggregate relation}, 
$u$ is a term and
$E$ is a (possibly infinite) collection of aggregate elements,
which are syntactically separated by \quo{;}.
\nop{\footnote{%
We allow for infinite collections of aggregate elements because the 
semantics in Section~\ref{sec:semantics} is based on ground instantiation,
which may map some non-ground aggregate element to infinitely many ground instances.
The semantics of \nf{Abstract Gringo} \cite{gehakalisc15a}
handles such cases by means of infinitary propositional formulas, while 
the \nf{Abstract Gringo} language avoids infinite collections of aggregate elements
in the input.
As shown in~\cite{DBLP:journals/tplp/HarrisonL19},
the semantics by ground instantiation or infinitary propositional formulas,
respectively, are equivalent on the common subset of
\nf{Abstract Gringo} and \langname.
Moreover, we note that the restrictions to \langname programs
claimed in Section~\ref{sec:aspcompnotes} require the existence of
a finite equivalent ground instantiation for each input, so that infinite collections of
aggregate elements do not show up in practice.}}%
%
Given an aggregate atom~$a$,
the expressions $a$ and $\naf a$ are {\em aggregate literals}.
%
In the following, we write {\em atom} (resp., {\em literal}) without further qualification
to refer to some classical, built-in or aggregate atom (resp., naf- or aggregate literal).

We here allow for infinite collections of aggregate elements because the
semantics in Section~\ref{sec:semantics} is based on ground instantiation,
which may map some non-ground aggregate element to infinitely many ground instances.
The semantics of \nf{Abstract Gringo} \cite{gehakalisc15a}
handles such cases by means of infinitary propositional formulas, while
the \nf{Abstract Gringo} language avoids infinite collections of aggregate elements
in the input.
As shown in \cite{DBLP:journals/tplp/HarrisonL19},
the semantics by ground instantiation or infinitary propositional formulas,
respectively, are equivalent on the common subset of
\nf{Abstract Gringo} and \langname.
Moreover, we note that the restrictions to \langname programs
claimed in Section~\ref{sec:aspcompnotes} require the existence of
a finite equivalent ground instantiation for each input, so that infinite collections of
aggregate elements do not show up in practice.

\paragraph{\bf Rules.}
\begin{nop}{A \emph{body element} is either a literal or an aggregate literal.
}\end{nop}

A {\em rule} has form
\[ h_1 \ | \ \dots \ | \ h_m \leftarrow b_1, \dots, b_n. \]
where $h_1, \dots, h_m$ are classical atoms
and $b_1, \dots, b_n$ are literals for $m\geq0$ and $n\geq0$.
When $n=0$, the rule is called a {\em fact}.
When $m=0$, the rule is referred to as a {\em constraint}.

\begin{nop}{
A rule is called {\em integrity constraint} if $m$ is zero, and
fact if it is ground, $m$ is one and $n$ is zero.
}\end{nop}

\begin{nop}{
\[
a_1 \vee \dots \vee a_n \leftarrow b_1, \dots , b_k, o_1, \dots, o_l, \naf n_1, ..., \naf n_m.
\]
where $n,k,m,l \geq 0$, and at least one of $n$,$k$ and $m$ is greater than $0$.
$a_1,\dots, a_n$, $b_1,\dots,b_k$, and $n_1,\dots,n_m$ are {\em classical literals}, while $o_1, \dots, o_l$ are
{\em builtin atoms}.
$a_1 \vee \dots \vee\, a_n$ constitutes the {\em head} of $r$, while $b_1, \dots , b_k, \naf n_1, ..., \naf n_m$ is the {\em body} of $r$.
As usual, whenever $k=m=l=0$, we omit the \quo{$\leftarrow$} sign. We call $r$ a {\em fact} if $n=1, k=m=0$ or a
{\em constraint} if $n=0$.

}\end{nop}





\paragraph{\bf Weak Constraints.}
A {\em weak constraint} has form
\[ \colonsim b_1, \dots, b_n.\, [ w@l,t_1, \dots, t_m ] \]
where
$t_1, \dots, t_m$ are terms and
$b_1, \dots, b_n$ are literals for $m\geq0$ and $n\geq0$;
$w$ and $l$ are terms standing for a {\em weight} and a {\em level}.
Writing the part ``$@l$'' can optionally be omitted if $l=0$;
that is, a weak constraint has level~$0$ unless specified otherwise.

\paragraph{\bf Queries.}
A {\em query} $Q$ has form $a?$, where $a$ is a classical atom.

\paragraph{\bf Programs.}
An {\em \langname program} 
is a set of rules and
weak constraints,
possibly accompanied by a (single) query.\footnote{%
Unions of conjunctive queries (and more)
can be expressed by including appropriate rules in a program.%
}
A program
(rule, weak constraint, query, literal, aggregate element, etc.)
is {\em ground} if it
contains no variables.


\section{Semantics}
\label{sec:semantics}

We herein give the full model-theoretic semantics of \langname.
As for non-ground programs, the semantics  extends the traditional notion of Herbrand interpretation,
taking care of the fact that {\em all} integers are part of the Herbrand universe.
The semantics of propositional programs is based on \cite{gelf-lifs-91},
extended to aggregates according to \cite{fabe-etal-2004-jelia,fabe-etal-2011-aij}.
Choice atoms \cite{simo-etal-2002} are treated in terms of the reduction given in
Section~\ref{sec:short}.

We restrict the given semantics to programs containing non-recursive aggregates
(see Section~\ref{sec:aspcompnotes} for this and further restrictions to the family of admissible programs),
for which the general semantics presented herein
is in substantial agreement with 
a variety of proposals for adding aggregates to ASP
\cite{kemp-stuc-91,vang-92,osor-jaya-99,ross-sagi-1997,dene-etal-2001,gelf-2002,simo-etal-2002,dell-etal-2003b,pelo-trus-2004,pelo-etal-2004a,ferr-2005-lpnmr,pelo-etal-2007-tplp}.

\paragraph{\bf Herbrand Interpretation.}

Given a program~$P$, the {\em Herbrand universe} of~$P$, denoted by \UP, consists of all integers
and (ground) terms constructible from constants and functors appearing in~$P$.
The {\em Herbrand base} of~$P$, denoted by \BP,
is the set of all (ground) classical atoms that can be built by combining
predicate names appearing in~$P$ with terms from \UP as arguments.
A (Herbrand) {\em interpretation} $I$ for~$P$ is a subset of \BP. 

\begin{nop}{
\subsection{Global and Local Variables.}
A {\em local} variable of a rule or weak constraint~$r$ is a variable appearing in aggregate elements only;
all other variables are {\em global} in~$r$. 
}\end{nop}

\paragraph{\bf Ground Instantiation.}

A {\em substitution} $\sigma$ is a mapping from a set~$V$ of variables
to the Herbrand universe~\UP of a given program~$P$.
For 
some object~$O$ (rule, weak constraint, query, literal, aggregate element, etc.),
we denote by $O\sigma$ the object obtained by replacing each
occurrence of a variable~$v\in V$ by~$\sigma(v)$ in~$O$.

A variable is {\em global} in a rule, weak constraint or query~$r$
if it appears outside of aggregate elements in~$r$.
A substitution from the set of global variables in~$r$ is a {\em global substitution for}~$r$;
a substitution from the set of 
variables in an aggregate element~$e$ is a (local) {\em substitution for}~$e$.
A global substitution~$\sigma$ for~$r$ (or substitution~$\sigma$ for~$e$) is
{\em well-formed} \comment{kali: move from {\em consistent} to {\em well-formed},
according to last conference call and given the current loss of proposals -- please
pay attention to any other occurrence within the paper}
if the  arithmetic  evaluation, performed in the standard way,
of any arithmetic subterm
($-(t)$ or $(t \diamond u)$ with $\diamond \in \{ \quo{+},\quo{-},\linebreak[1]\quo{*}, \quo{/} \}$)
appearing outside of aggregate elements in~$r\sigma$ (or appearing in~$e\sigma$)
is well-defined.

Given a collection~$E$ 
of aggregate elements, 
the {\em instantiation} of~$E$ 
is the following set of 
aggregate elements:
%
\[\mathrm{inst}(E) = \mbox{$\bigcup$}_{e\in E}\{ e\sigma \mid
\sigma \mbox{ is a well-formed 
substitution for }e \} \]
A {\em ground instance} of a rule, weak constraint or query~$r$ is obtained in two steps:
(1) a well-formed global substitution~$\sigma$ for~$r$ is applied to~$r$;
(2) for every aggregate atom 
$\aggr{aggr}\ E \prec\nolinebreak u$
appearing in~$r\sigma$, $E$ 
is replaced by 
$\mathrm{inst}(E)$. 

The {\em arithmetic evaluation} of a ground instance~$r$ of some rule, weak constraint or query
is obtained by replacing any maximal arithmetic subterm
appearing in~$r$ 
by its integer value,
which is calculated in the standard way.
\footnote{Note that the outcomes of  arithmetic  evaluation are well-defined
          relative to well-formed substitutions.}
%
%
The {\em ground instantiation} of a program~$P$, 
denoted by \GP,
is the set of  arithmetically  evaluated ground instances of rules and
weak constraints in~$P$.


\begin{nop}{
\subsection{Evaluation of Terms.}
The evaluation of some (ground) term is obtained by
replacing its outermost arithmetic subterms
(of form $-(t)$ or $(t \diamond u)$ with $\diamond \in \{ \quo{+},\quo{-},\quo{*}, \quo{/} \}$)
by their integer values calculated in the standard way.
If the integer value of some arithmetic subterm is undefined
(e.g.\ due to division by $0$),
the evaluation of any encompassing term is undefined as well.
}\end{nop}

\paragraph{\bf Term Ordering and Satisfaction of Naf-Literals.}

A classical atom $a\in\BP$ is {\em true} w.r.t.\ a 
interpretation~$I\subseteq\BP$ if $a\in I$. 
A Naf-Literal of the form $\naf\ a$, where $a$ is a classical atom, is {\em true} w.r.t. $I$ if $a \notin I$, and it is false otherwise.

To determine whether a built-in atom $t \prec u$ (with
${\prec} \in \{ \quo{<}, \quo{\leq}, \quo{=}, \quo{\neq},\allowbreak \quo{>}, \quo{\geq} \}$)
holds,
we rely on a    total order $\preceq$ on terms in \UP defined as follows:
\begin{itemize}
\item $t \preceq u$ for integers $t$ and $u$ if $t\leq u$;
\item $t \preceq u$ for any integer~$t$ and any symbolic constant~$u$;
\item $t \preceq u$ for symbolic constants $t$ and $u$ if $t$ is lexicographically smaller than or equal to~$u$;
\item $t \preceq u$ for any symbolic constant~$t$ and any string constant~$u$;
\item $t \preceq u$ for string constants $t$ and $u$ if $t$ is lexicographically smaller than or equal to~$u$;
\item $t \preceq u$ for any string constant~$t$ and any functional term~$u$;
\item $t \preceq u$ for functional terms $t=f(t_1,\dots,t_m)$ and $u=g(u_1,\dots,u_n)$ if
  \begin{itemize}
    \item $m<n$ (the arity of~$t$ is smaller than the arity of~$u$),
    \item $m\leq n$ and $g\npreceq f$ (the functor of~$t$ is smaller than the one of~$u$, while arities coincide) or
    \item $m\leq n$, $f\preceq g$ and, for any $1\leq j\leq m$ such that $t_j\npreceq u_j$,
          there is some $1\leq i<j$ such that $u_i\npreceq t_i$
          (the tuple of arguments of~$t$ is smaller than or equal to the arguments of~$u$).
  \end{itemize}
\end{itemize}
Then, $t \prec u$ is {\em true} w.r.t.~$I$ if
$t \preceq u$ for ${\prec} = \quo{\leq}$;
$u \preceq t$ for ${\prec} = \quo{\geq}$;
$t \preceq u$ and $u \npreceq t$ for ${\prec} = \quo{<}$;
$u \preceq t$ and $t \npreceq u$ for ${\prec} = \quo{>}$;
$t \preceq u$ and $u \preceq t$ for ${\prec} = \quo{=}$;
$t \npreceq u$ or $u \npreceq t$ for ${\prec} = \quo{\neq}$.
A  positive  naf-literal $a$ is {\em true} w.r.t.~$I$
if $a$ is a classical or built-in atom that is {\em true} w.r.t.~$I$;
otherwise, $a$ is {\em false} w.r.t.~$I$.
A  negative  naf-literal $\naf a$ is {\em true} (or {\em false}) w.r.t.~$I$
if $a$ is {\em false} (or {\em true}) w.r.t.~$I$.

\begin{nop}{
The truth value of a (ground) built-in of form $x \diamond y$ is determined in the standard way
by evaluating terms $x$ and $y$ and then checking the corresponding relation $\diamond$.
If the relation $\diamond$ is not defined for the arguments,
the truth value of the built-in is undefined.

The relations $=$ and $\neq$ are defined for all ground terms;
all other relations are only defined over integer (standard ordering) and string (lexicographic ordering see Section \ref{sec:gram}) domains .

\comment{RK: We could also establish an ordering say integer $<$ symbolic constant $<$ string constant $<$ function symbol.}
\comment{RK: Actually, we already made use of such orderings in encodings (especially in min/max aggregates).}
\comment{RK: Changed from false to undefined; this should not be relied upon.}

\subsection{Satisfaction of Literals.}
A positive naf-literal
$l = a$ (resp., a naf-literal $l = \naf\ a$), for $a$ a predicate atom, is true w.r.t.
$I$ if $a \in I$ (resp., $a \notin I$); it is false otherwise.
}\end{nop}

\paragraph{\bf Satisfaction of Aggregate Literals.}

An {\em aggregate function} is a mapping from
sets of tuples of terms to terms, $+\infty$ or $-\infty$.
The aggregate functions associated with
aggregate function names introduced in Section~\ref{sec:language}
map a set~$T$ of tuples of terms to a term, $+\infty$ or $-\infty$ as follows:\footnote{%
The special cases in which
$\aggr{aggr}(T)=+\infty$, $\aggr{aggr}(T)=-\infty$ or
$\aggr{sum}(T) = 0$ for an infinite set
$\{(t_1, \dots, t_m)\in T \mid t_1$ is a non-zero integer$\}$
are adopted from \nf{Abstract Gringo} \cite{gehakalisc15a}.}
\begin{itemize}
\item $\aggr{count}(T) =
\left\{
\begin{array}{@{}l@{\text{ \ if }}l@{}}
|T| & T \text{ is finite}
\\
+\infty & T \text{ is infinite;}
\end{array}
\right.$
\item $\aggr{sum}(T) =
\left\{
\begin{array}{@{}l@{\text{ \ if }}l@{}}
\sum_{\text{$(t_1, \dots, t_m)\in T$, $t_1$ is an integer}}t_1
&
\{(t_1, \dots, t_m)\in T \mid t_1 \text{ is a non-zero integer}\}
\text{ is finite}
\\
0
&
\{(t_1, \dots, t_m)\in T \mid t_1 \text{ is a non-zero integer}\}
\text{ is infinite;}
\end{array}
\right.$
\item $\aggr{max}(T) =
\left\{
\begin{array}{@{}l@{\text{ \ if }}l@{}}
\max \{t_1 \mid (t_1, \dots, t_m)\in T\}
&
T\neq\emptyset \text{ is finite}
\\
+\infty
&
T \text{ is infinite}
\\
-\infty
&
T=\emptyset\text{;}
\end{array}
\right.$
\item $\aggr{min}(T) =
\left\{
\begin{array}{@{}l@{\text{ \ if }}l@{}}
\min \{t_1 \mid (t_1, \dots, t_m)\in T\}
&
T\neq\emptyset \text{ is finite}
\\
-\infty
&
T \text{ is infinite}
\\
+\infty
&
T=\emptyset\text{.}
\end{array}
\right.$
\end{itemize}
The terms selected by $\aggr{max}(T)$ and $\aggr{min}(T)$
for finite sets $T\neq\emptyset$
are determined relative to the total order~$\preceq$ on terms in \UP.
In the special cases that
$\aggr{aggr}(T)=+\infty$ or $\aggr{aggr}(T)=-\infty$,
we adopt the convention that
$-\infty \preceq u$ and $u \preceq +\infty$
for every term $u\in \UP$.
\begin{nop}{
\comment{MG: Can this convention be agreed upon?\\
             (In \texttt{gringo}, $\aggr{max}([])=\aggr{infimum}$ and
              $\aggr{min}([])=\aggr{supremum}$, where $\aggr{infimum}$
              and $\aggr{supremum}$ are constants standing for the
              greatest lower and least upper bound of $\preceq$.)}
}\end{nop}%
An expression
$\aggr{aggr}(T)\prec u$ is {\em true} (or {\em false})
for
$\aggr{aggr} \in \{ \quo{\aggr{count}}, \quo{\aggr{sum}}, \quo{\aggr{max}},\allowbreak \quo{\aggr{min}} \}$,
an aggregate relation
${\prec} \in \{ \quo{<}, \quo{\leq}, \quo{=}, \quo{\neq},\allowbreak \quo{>}, \quo{\geq} \}$
and a term~$u$
if $\aggr{aggr}(T)\prec u$ is {\em true} (or {\em false})
according to the corresponding definition for built-in atoms,
given previously, extended to the values $+\infty$ and $-\infty$ for $\aggr{aggr}(T)$.

An interpretation $I\subseteq\BP$ maps
a collection~$E$ of aggregate elements
to the following set of tuples of terms:
\[
\mathrm{eval}(E,I)
=
\{(t_1, \dots, t_m) \mid
t_1, \dots, t_m : l_1, \dots, l_n
\text{ \ occurs in $E$ and $l_1, \dots, l_n$ are {\em true} w.r.t.\ $I$}
\}
\]
A  positive  aggregate literal 
$a=\aggr{aggr}\ E \prec u$
is {\em true} (or {\em false}) w.r.t.~$I$ if
$\aggr{aggr}(\mathrm{eval}(E,I))\prec u$
is {\em true} (or {\em false}) w.r.t.~$I$;
$\naf a$ is {\em true} (or {\em false}) w.r.t.~$I$
if $a$ is {\em false} (or {\em true}) w.r.t.~$I$.%

\begin{nop}{
We associate to each aggregate function name $\aggr[n]{f}$ a corresponding {\em aggregate function} $f$ mapping multisets to integer values.
For a sequence of ground aggregate elements $E$, let $\Pi(E)$ be its corresponding set.

Let $I$ be an interpretation.
The valuation~$I(S)$ of~$S$ w.r.t.~$I$ is the multiset of the first constant of the elements in~$S$ whose literals are all true w.r.t.~$I$.
More precisely, let~$I(S)$ denote the multiset
\[ [ t_1 \mid t_1,\dots,t_m \in \{ \vec{t} \mid \vec{t} \Colon \vec{l} \in S \text{ and each $l\in\vec{l}$ is satisfied} \} ] \]

The {\em valuation}~$V(I,S)$ of a ground aggregate atom~$A = \aggr{f} \{ S \} \prec t$ is defined as~$f(\Pi(I(S)))$.
\comment{RK: I removed multiset aggregates. I think that if we want to support this, we should provide it as syntactic shortcut.}
\comment{RK: I adjusted the definition slightly to properly emphasize what goes into the multiset.}

The {\em truth value w.r.t.~$I$} of a ground aggregate atom of the form~$A = \aggr{f} \{ S \} \prec t$ is determined the same way as for the build in $V(I,S) \prec t$.
The truth value of a negative aggregate literal is determined analogously with the truth value flipped.

\subsection{Available Aggregate Functions.}
The available aggregate functions in \langname are defined as:
\begin{itemize}
  \item $\mathit{count}(S) = |S|$;
  \item $\mathit{sum}(S) = \Sigma_{s \in S} s$ where $\mathit{sum}$ is defined only for multisets of integers;
  \item $\mathit{max}(S) = \max S$;
  \item $\mathit{min}(S) = \min S$.
\end{itemize}
Both $\mathit{min}$ and $\mathit{max}$ are defined over homogeneous sets of integers, strings or quoted strings.
For the latter two cases it is considered the total partial order enforced by the program character encoding (see Section \ref{sec:gram} for details).
If the multiset $S$ is not in the domain of an aggregate function $f$, the result of the function is undefined.
\comment{RK: Again, I changed false to undefined because having aggregate $A$ and $\naf A$ evaluate to false is strange.}
}\end{nop}

\paragraph{\bf Answer Sets.}

Given a program~$P$ and a (consistent) interpretation $I\subseteq\BP$,
a rule $h_1\ |\ \dots\ |\ h_m \leftarrow b_1, \dots, b_n.$ in $\GP$
is {\em satisfied} w.r.t.~$I$ if
some $h\in\{h_1, \dots, h_m\}$ is {\em true} w.r.t.~$I$ when
$b_1, \dots, b_n$ are {\em true} w.r.t.~$I$;
$I$ is a {\em model} of~$P$
if every rule in $\GP$ is satisfied w.r.t.~$I$.
The {\em reduct} of $P$ w.r.t.~$I$, denoted by~$P^I$,
consists of the rules $h_1\ |\ \dots\ |\ h_m \leftarrow b_1, \dots, b_n.$ in $\GP$
such that $b_1, \dots, b_n$ are {\em true} w.r.t.~$I$;
$I$ is an {\em answer set} of~$P$ if $I$ is a $\subseteq$-minimal model
of $P^I$. In other words, an answer set~$I$ of~$P$ is a model of~$P$
such that no proper subset of~$I$ is a model of~$P^I$.

The semantics of $P$ is given by the collection of its answer sets,
denoted by $\AS(P)$.

\paragraph{\bf Optimal Answer Sets.}

To select optimal members of $\AS(P)$, we map
an interpretation~$I$ for~$P$ to a set of tuples as follows:
%
\begin{align*}
\mathrm{weak}(P,I) =
\{ & (w@l,t_1, \dots, t_m) \mid {}
\\ &
  \colonsim b_1, \dots, b_n.\, [ w@l,t_1, \dots, t_m ]
  \text{ \ occurs in \GP
  and $b_1, \dots, b_n$ are {\em true} w.r.t.~$I$}
\}
\end{align*}
For any integer~$l$,
let
\[
P^I_{\!l}
=
\left\{
\begin{array}{@{}l@{\text{ \ if }}l@{}}
\mbox{$\sum$}_{\text{$(w@l,t_1, \dots, t_m)\in \mathrm{weak}(P,I)$, $w$ is an integer}}w
&
\{(w@l,t_1, \dots, t_m)\in \mathrm{weak}(P,I) \mid w \text{ is a non-zero integer}\}
\\
\multicolumn{1}{@{}l@{}}{} &
\multicolumn{1}{@{}l@{}}{\text{is finite}}
\\
0
&
\{(w@l,t_1, \dots, t_m)\in \mathrm{weak}(P,I) \mid w \text{ is a non-zero integer}\}
\\
\multicolumn{1}{@{}l@{}}{} &
\multicolumn{1}{@{}l@{}}{\text{is infinite}}
\end{array}
\right.
\]
denote the sum of integers~$w$ over 
tuples with
$w@l$ in $\mathrm{weak}(P,I)$.
Then, an answer set~$I\in\AS(P)$ is {\em dominated}
by 
$I'\in\AS(P)$ if there is some integer~$l$ such that
$P^{I'\!}_{\!l}<P^I_{\!l}$ and
$P^{I'\!}_{\!l'}=P^I_{\!l'}$ for all integers $l'>l$.
An answer set~$I\in\AS(P)$ is {\em optimal}
if there is no $I'\in\AS(P)$ 
such that $I$ is dominated by~$I'$.
Note that~$P$ has some (and possibly more than one)
optimal answer sets if $\AS(P)\neq\emptyset$.

\paragraph{\bf Queries.}
%
%
%
Given a ground query $Q = q?$ of a program $P$,
$Q$ is true if $q\in I$ for all $I\in\AS(P)$. 
Otherwise, $Q$ is false. Note that, if $\AS(P) = \emptyset$, all queries
are true.
%
%
In presence of variables one is interested in substitutions that make the query true.
Given the non-ground query $Q = q(t_1,\dots,t_n)?$ of a program $P$, let
$\Ans(Q,P)$ be the set of all substitutions $\sigma$ for
$Q$ such that $Q\sigma$ is true. The set $\Ans(Q,P)$ constitutes
the set of answers to~$Q$.
Note that, if $\AS(P) = \emptyset$, $\Ans(Q,P)$ contains all
possible substitutions for~$Q$.

Note that query answering, according to the definitions above, corresponds
to cautious (skeptical) reasoning as defined in, e.g., \cite{abit-etal-95}.


\begin{nop}{
\subsection{Satisfaction of Rules.}
Given a ground rule $r$, we say that $r$ is satisfied w.r.t. $I$
if some naf-literal appearing in the head of $r$ is true w.r.t. $I$ or some
naf-literal appearing in the body of $r$ is false w.r.t. $I$. Given a ground
program $P$, we say that $I$ is a {\em model} \ of $P$, iff all
rules in \GP are satisfied w.r.t. $I$. A model $M$ is {\em
minimal} if there is no model $N$ for $P$ such that $N \subset M$.


\subsection{Generalized Gelfond-Lifschitz Reduct and Answer Sets\cite{fabe-etal-2004-jelia,fabe-etal-2011-aij}.}
\label{subsec:answersets}

We are given a ground program $P$ without weak constraints;
let $I$ be an interpretation,
let $P^I$ denote the transformed program obtained from $P$ by deleting all rules in which a body naf-literal is false w.r.t.\ $I$.
$I$ is an {\em answer set} of a program $\p$ if it is a minimal model of $P^I$.
Let $AS(P)$ be the set of answer sets of $P$.

An interpretation $I$ is an answer set of $P$ if there exists an interpretation $I' = I \cup Aux$
such that $I \in AS(gr(P'))$ where $Aux$ is an interpretation defined over predicate names appearing in $P'$ but not in $P$.

Given a logic program with weak constraints;
an {\em optimal answer set} is an answer set of the corresponding logic program with all weak constraints removed
that maximizes (lexicographically) the sums of the weights of satisfied weak constraints grouped and ordered by level. 
}\end{nop}


\section{Syntactic Shortcuts}
\label{sec:short}

This section specifies additional constructs
by reduction to the language introduced in Section~\ref{sec:language}.

\paragraph{\bf Anonymous Variables.}

An {\em anonymous variable} in a rule, weak constraint or query is denoted by
\quo{$\_$} (character underscore).
Each occurrence of \quo{$\_$} stands
for a fresh variable in the respective context 
(i.e.,
different occurrences of anonymous variables represent distinct variables).

\paragraph{\bf Choice Rules.}

A {\em choice element} has form
\[ a : l_1, \dots, l_k \] 
where $a$ is a classical atom and $l_1, \dots, l_k$ are naf-literals for $k\geq 0$.

A {\em choice atom} has form
\[ C \prec u \]
where $C$ is a collection of choice elements,
which are syntactically separated by \quo{;},
$\prec$ is an aggregate relation (see Section \ref{sec:language}) and
$u$ is a term. The part ``$\prec u$'' can optionally be
omitted if ${\prec}$ is 
$\quo{\geq}$ and $u=0$.

A {\em choice rule} has form
\[ C \prec u \leftarrow b_1, \dots, b_n. \]
where $C \prec u$ 
is a choice atom and
$b_1, \dots, b_n$ are literals for $n\geq0$.

Intuitively, a choice rule means that, if the body of the rule is {\em true}, 
an arbitrary subset of the classical atoms $a$ 
such that $l_1, \dots, l_k$ are {\em true}
can be chosen
as {\em true} in order to comply with the 
aggregate relation~$\prec$ between $C$ and~$u$.
In the following,
this intuition is captured 
by means of a proper mapping of choice rules
to rules without choice atoms (in the head).

For any predicate atom $q=p(t_1,\dots,t_n)$,
let
$\widehat{q}=\hat{p}(1,t_1,\dots,t_n)$ and
$\widehat{\neg q}=\hat{p}(0,t_1,\dots,t_n)$,
where $\hat{p}\neq p$ is an (arbitrary) predicate and function name
that is uniquely associated with~$p$, 
and the first argument (which can be $1$ or $0$) indicates
the \quo{polarity} $q$ or $\neg q$, respectively.%
%
\footnote{It is assumed that fresh predicate and function names are outside of possible program signatures and cannot be referred to within user
input.}

Then, a choice rule 
stands for the rules
\begin{align*}
  a\ |\ \widehat{a} 
& {} \leftarrow b_1, \dots, b_n, 
                l_1, \dots, l_k.
\end{align*}
for each choice element $a : l_1, \dots, l_k$ in~$C$ 
along with the 
constraint
\begin{align*}
& {} \leftarrow b_1, \dots, b_n, 
                \naf \aggr{count}
                \{\,
                   \widehat{a} : a,l_1, \dots, l_k
                   \mid
                   (a : l_1, \dots, l_k) \in C
                \} 
                \prec u.
\end{align*}
%
The first group of rules expresses that
the classical atom~$a$ 
in a choice element $a : l_1, \dots, l_k$
can be chosen as {\em true} (or {\em false}) if
$b_1, \dots, b_n$ and 
$l_1, \dots, l_k$ 
are {\em true}.
This \quo{generates} all subsets of the atoms in choice elements.
On the other hand,
the second rule, which is an integrity constraint, requires the condition
$C \prec u$ 
to hold
if $b_1, \dots, b_n$ are {\em true}.%
\footnote{%
In disjunctive heads of rules of the first form,
an occurrence of $\widehat{a}$ 
denotes an (auxiliary) {\em atom} that is
linked to the original atom~$a$. 
Given the relationship between $a$ 
and~$\widehat{a}$, 
the latter is reused 
as a {\em term} 
in the body of a rule of the second form.
That is, we overload the notation $\widehat{a}$ 
by letting it stand both for an atom (in disjunctive heads) and
a term (in $\aggr{count}$ aggregates).}

For illustration, consider the choice rule
\[
\{ p(a) : q(2); \neg p(a) : q(3) \} \leq 1 \leftarrow q(1).
\]
Using the fresh predicate and function name~$\hat{p}$,
the  choice rule is mapped to three rules as follows:
\begin{align*}
  p(a)\ |\ \hat{p}(1,a)
& {} \leftarrow q(1), q(2).
\\
  \neg p(a)\ |\ \hat{p}(0,a)
& {} \leftarrow q(1), q(3).
\\
& {} \leftarrow q(1), \naf \aggr{count} \{ \hat{p}(1,a) : p(a), q(2);
                                           \hat{p}(0,a) : \neg p(a), q(3) \}  \leq 1.
\end{align*}
Note that the three rules are satisfied w.r.t.\ an interpretation~$I$
such that 
$\{q(1),\linebreak[1]q(2),\linebreak[1]q(3),\linebreak[1]\hat{p}(1,a),\linebreak[1]\hat{p}(0,a)\}\subseteq I$
and
$\{p(a),\neg p(a)\}\cap I=\emptyset$.
In fact, when $q(1)$, $q(2)$, and $q(3)$ are {\em true}, 
the choice of none or one of the atoms $p(a)$ and $\neg p(a)$
complies with the aggregate relation $\quo{\leq}$ to~$1$.

\begin{nop}{
\begin{align*}
  \beta & \leftarrow B \\
        & \leftarrow \beta, \naf \aggr{count} \{ p_1,n_1,\vec{t}_1 : a_1, \alpha_1; \dots; p_n,o_n,\vec{t}_n : a_n, \alpha_n \} \prec t
\end{align*}
and the rules
\begin{align*}
  \alpha_i & \leftarrow l^i_1, \dots, l^i_{n_i}, \beta \\
  a_i; \bar{a}_i & \leftarrow \alpha_i
\end{align*}
for each $e_i$ of form $ a_i : l^i_1, \dots, l^i_{n_i} $
where $1 \leq i \leq n$ and $p_i$, $o_i$ and $\vec{t}_i$ are the predicate symbol, arity and arguments associated with $a_i$.
Furthermore, $\beta$ is an auxiliary predicate atom that has all the global variables in $B$ as arguments
and each $\alpha_i$ is an auxiliary predicate atom that has all the global variables in $l^i_1, \dots, l^i_{n_i}, \beta$ as arguments.%
\footnote{Here this translation is given as a reference.
Take note that
answer sets are defined on a vocabulary excluding auxiliary names, see section \ref{subsec:answersets}.
This allows implementors to possibly use other strongly equivalent
(relatively to the vocabulary without aux predicates) reductions/evaluation techniques.}
}\end{nop}

\paragraph{\bf Aggregate Relations.}

An aggregate or choice atom
\[ \aggr{aggr}\ E \prec u 
   \text{ \ \ or \ \ }
                C \prec u \] 
may be written as
\[ u \prec^{-1} \aggr{aggr}\ E 
   \text{ \ \ or \ \ }
   u \prec^{-1} C \] 
where  $\quo{<}^{-1} = \quo{>}$; \ $\quo{\leq}^{-1}  =  \quo{\geq}$; \ $\quo{=}^{-1} = \quo{=}$; \
$\quo{\neq}^{-1} = \quo{\neq}$; \ $\quo{>}^{-1} = \quo{<}$; \ $\quo{\geq}^{-1} = \quo{\leq}$.

The left and right notation of aggregate relations
may be combined in expressions as follows:
\[ u_1 \prec_1 \aggr{aggr}\ E \prec_2 u_2 
   \text{ \ \ or \ \ }
   u_1 \prec_1 C \prec_2 u_2 \] 
Such expressions are mapped to available constructs
according to the following transformations:
\begin{description}
\item[%
\(
\diamond\ u_1 \prec_1 C 
              \prec_2 u_2 \leftarrow b_1, \dots, b_n.
\)]
stands for
\begin{align*}
u_1 \prec_1 C 
              & {} \leftarrow b_1, \dots, b_n.
\\
C \prec_2 u_2 & {} \leftarrow b_1, \dots, b_n.
\end{align*}
\item[%
\(
\diamond\ h_1\ |\ \dots\ |\ h_k \leftarrow b_1, \dots,b_{i-1},
                          u_1 \prec_1 \aggr{aggr}\ E 
                              \prec_2 u_2
                          ,b_{i+1},\dots, b_n.
\)]
stands for
\begin{align*}
h_1\ |\ \dots\ |\ h_k & {} \leftarrow b_1, \dots,b_{i-1},
                               u_1 \prec_1 \aggr{aggr}\ E, 
                               \aggr{aggr}\ E \prec_2 u_2  
                               ,b_{i+1},\dots, b_n.
\end{align*}
\item[%
\(
\diamond\ h_1\ |\ \dots\ |\ h_k \leftarrow b_1, \dots,b_{i-1},
                          \naf u_1 \prec_1 \aggr{aggr}\ E \prec_2 u_2 
                          ,b_{i+1},\dots, b_n.
\)]
stands for
\begin{align*}
h_1\ |\ \dots\ |\ h_k & {} \leftarrow b_1, \dots,b_{i-1},
                               \naf u_1 \prec_1 \aggr{aggr}\ E 
                               ,b_{i+1},\dots, b_n.
\\
h_1\ |\ \dots\ |\ h_k & {} \leftarrow b_1, \dots,b_{i-1},
                               \naf \aggr{aggr}\ E \prec_2 u_2 
                               ,b_{i+1},\dots, b_n.
\end{align*}
\item[%
\(
\diamond\ \colonsim b_1, \dots,b_{i-1},
          u_1 \prec_1 \aggr{aggr}\ E \prec_2 u_2 
         ,b_{i+1},\dots, b_n.\, {[} w@l,t_1, \dots, t_k {]}
\)]
stands for
\begin{align*}
\hspace*{-3mm} &
\colonsim b_1, \dots,b_{i-1},
          u_1 \prec_1 \aggr{aggr}\ E, 
          \aggr{aggr}\ E \prec_2 u_2  
         ,b_{i+1},\dots, b_n.\, [ w@l,t_1, \dots, t_k ]
\end{align*}
\item[%
\(
\diamond\ \colonsim b_1, \dots,b_{i-1},
          \naf u_1 \prec_1 \aggr{aggr}\ E \prec_2 u_2 
         ,b_{i+1},\dots, b_n.\, {[} w@l,t_1, \dots, t_k {]}
\)]
stands for
\begin{align*}
&
\colonsim b_1, \dots,b_{i-1},
          \naf u_1 \prec_1 \aggr{aggr}\ E 
         ,b_{i+1},\dots, b_n.\, [ w@l,t_1, \dots, t_k ]
\\
&
\colonsim b_1, \dots,b_{i-1},
          \naf \aggr{aggr}\ E \prec_2 u_2 
         ,b_{i+1},\dots, b_n.\, [ w@l,t_1, \dots, t_k ]
\end{align*}
\end{description}

\begin{nop}{
A positive aggregate literal with \emph{two bounds} occurring in the body of a rule or weak constraint
\[ R(t_1 \prec_1 \aggr{aggr}\, S \prec_2 t_2) \]
is a shortcut for
\[ R(\aggr{aggr}\, S \prec_1^{-1} t_1, \aggr{aggr}\, S \prec_2 t_2) \]
and a negative aggregate literal with \emph{two bounds} occurring in the body of a rule or weak constraint
\[ R(\naf t_1 \prec_1 \aggr{aggr}\, S \prec_2 t_2) \]
is a shortcut for
\[ R(\naf \aggr{aggr}\, S \prec_1^{-1} t_1) \]
\[ R(\naf \aggr{aggr}\, S \prec_2 t_2) \]
and a choice atom with \emph{two bounds} in a rule head
\[ t_1 \prec_1 S \prec_2 t_2 \leftarrow B \]
is a shortcut for
\[ S \prec_1^{-1} t_1 \leftarrow B \]
\[ S \prec_2 t_2 \leftarrow B \]
where $\prec_1^{-1}$ is the inverse aggregate relation of $\prec_1$.
A choice atom $S$ \emph{without bounds} is a shortcut for $S \geq 0$.
\comment{RK: I patched a bit here to keep the rewriting in line with the aggregate atom definition.
(Admittedly, the above is a bit ugly but should work.)
After some thoughts, I think we do not have to restrict the use of relations here.}
}\end{nop}

\begin{nop}{
\subsection{Optimize Statements.}
\label{subsec:minimize}

An {\em optimize statement} has form
\[
\aggr{opt} 
\{ 
   w_1@l_1,t_{1_1}, \dots, t_{m_1} : b_{1_1}, \dots, b_{n_1}; 
   \dots ; 
   w_k@l_k,t_{1_k}, \dots, t_{m_k} : b_{1_k}, \dots, b_{n_k}
\}.
\]
where
$\aggr{opt} \in \{ \quo{\aggr{minimize}}, \quo{\aggr{maximize}} \}$,
$w_i,l_i,t_{1_i}, \dots, t_{m_i}$ are terms and
$b_{1_i}, \dots, b_{n_i}$ are naf-literals
for $k\geq 0$, $1\leq i\leq k$, $m_i\geq 0$ and $n_i\geq 0$.
Similar to weak constraints (cf.\ Section~\ref{subsec:weak}),
$w_i$ and $l_i$ stand for a {\em weight} and a {\em level}, and
writing ``$@l_i$'' can optionally be omitted if $l_i=0$.

An optimize statement stands for the weak constraints
\[
\colonsim b_{1_1}, \dots, b_{n_1}.\, [w_1'@l_1,t_{1_1}, \dots, t_{m_1}] 
\qquad\dots\qquad
\colonsim b_{1_k}, \dots, b_{n_k}.\, [w_k'@l_k,t_{1_k}, \dots, t_{m_k}]
\]
where $w_i'=w_i$ (or $w_i'=-w_i$) for $1\leq i\leq k$
if $\aggr{opt}=\quo{\aggr{minimize}}$ (or $\aggr{opt}=\quo{\aggr{maximize}}$).
}\end{nop}

\section{Using \langname in Practice -- Restrictions}
\label{sec:aspcompnotes}

To promote declarative programming
as well as practical system implementation,
\langname programs are supposed to comply with the restrictions
listed in this section.
This particularly applies to input programs starting from the System Track of the 4th Answer Set Programming Competition~\cite{CalimeriIR14}.

\paragraph{\bf Safety.}

Any rule, weak constraint or query is required to be safe;
to this end,
for a set $V$ of variables and
literals $b_1,\dots,b_n$,
we inductively (starting from an empty set of bound variables) define
$v\in V$ as {\em bound} by $b_1,\dots,b_n$ if
$v$ occurs outside of arithmetic terms
in some $b_i$ for $1\leq i\leq n$ such that $b_i$ is
\begin{itemize}
\item (i) a classical atom,
\item (ii) a built-in atom $t=u$ or $u=t$ and any member of~$V$ occurring in~$t$ is bound by $\{ b_1,\dots,b_n \} \setminus b_i$ or
\item (iii) an aggregate atom $\aggr{aggr}\,E=u$ and any member of~$V$ occurring in~$E$ is bound by $\{ b_1,\dots,b_n \} \setminus b_i$.
\end{itemize}
The entire set $V$ of variables
is {\em bound} by $b_1,\dots,b_n$ if each $v\in V$ is bound by $b_1,\dots,b_n$.

A rule, weak constraint or query~$r$ is {\em safe} if the
set~$V$ of global variables in~$r$ is bound by $b_1,\dots,b_n$
(taking a query~$r$ to be of form $b_1?$)
and, for each aggregate element $t_1, \dots, t_k : l_1, \dots, l_m$ in~$r$
with occurring variable set~$W$,
the set $W\setminus V$ of local
variables is bound by $l_1, \dots, l_m$.
For instance, the rule 

$$p(X,Y) \leftarrow q(X), \aggr{sum} \{ S,X : r(T,X), S=(2*T)-X \} = Y.$$ \\ 
is safe because all variables are bound by $q(X),r(T,X)$, while 

$$p(X,Y) \leftarrow q(X), \aggr{sum} \{ S,X : r(T,X), S+X=2*T \} = Y.$$  \\
is not safe because the expression $S+X=2*T$ does not respect condition (ii) above.

%

\paragraph{\bf Finiteness.}
Pragmatically, ASP programs solving real problems have a finite number of answer sets of finite size.
As an example, a program including $p(X+1) \leftarrow p(X).$ or $p(f(X)) \leftarrow p(X).$
along with a fact like $p(0).$ is not an admissible input in ASP Competitions.
There are pragmatic conditions that can be checked to ensure that a program admits finitely many answer sets (e.g.,~\cite{DBLP:journals/aicom/CalimeriCIL11}); in alternative, finiteness can be witnessed by providing a known maximum integer and maximum function nesting level per problem instance, which correctly limit the absolute values of integers as well as the depths of functional terms occurring as arguments in the atoms of answer sets.
The last option is the one adopted in ASP competitions since 2011.



\paragraph{\bf Aggregates.}

%
For the sake of an uncontroversial semantics, we require aggregates to be non-recursive.
To make this precise,
for any predicate atom $q=p(t_1,\dots,t_n)$,
let $q^v=p/n$ and $\neg q^v=\neg p/n$.
We further define the directed {\em predicate dependency graph} $D_P=(V,E)$ for a program~$P$
by
\begin{itemize}
\item the set~$V$ of vertices $a^v$
      for all classical atoms~$a$ appearing in~$P$ and
\item the set~$E$ of edges
      $(h_i^v,h_1^v),\dots,(h_i^v,h_m^v)$ and
      $(h_1^v,a^v),\dots,(h_m^v,a^v)$ for all rules
      \( h_1 \ | \ \dots \ | \ h_m \leftarrow b_1, \dots, b_n. \)
      in~$P$, $1\leq i\leq m$ and classical atoms~$a$ appearing in $b_1, \dots, b_n$.
\end{itemize}
The aggregates in~$P$ are {\em non-recursive} if,
for any classical atom $a$ appearing within aggregate elements
in a rule \( h_1 \ | \ \dots \ | \ h_m \leftarrow b_1, \dots, b_n. \)
in~$P$,
there is no path from $a^v$ to $h_i^v$ in $D_P$ for $1\leq i\leq m$.




\paragraph{\bf Predicate Arities.}

The arity of atoms sharing some predicate name is not assumed to be fixed.
However,
system implementers are encouraged to issue proper warning messages
if an input program includes
classical atoms with the same predicate name but different arities.

\paragraph{\bf Undefined Arithmetics.}
The semantics of \langname requires that substitutions that lead to undefined arithmetic subterms (and are thus not well-formed) are excluded by ground instantiation as specified in Section~\ref{sec:semantics}.
In practice, this condition is not easy to meet and implement for a number of technical reasons;
thus, it might cause problems to existing implementations, or even give rise to unexpected behaviors.

In order to avoid such complications, we require that a program~$P$ shall be invariant under undefined arithmetics;
that is, \GP is supposed to be equivalent to any ground program $P'$ obtainable from~$P$ by freely
replacing arithmetic subterms with undefined outcomes by arbitrary terms from \UP.
Intuitively, rules have to be written in such a way that the semantics of a   program does not change,
no matter the handling of substitutions that are not well-formed.

For instance, the 
program
\begin{align*}
& a(0).\\
& p \leftarrow a(X), \naf q(X/X).
\end{align*}
has the (single) answer set $\{a(0)\}$.
This program, however, is not invariant under undefined arithmetics.
Indeed, a vanilla grounder that skips arithmetic evaluation (in view
of no rule with atoms of predicate~$q$ in the head)
might produce 
the (simplified) ground rule $p \leftarrow a(0).$,
and this would result in the wrong answer set $\{a(0),p\}$.

In contrast to the previous program, 
\begin{align*}
& a(0).\\
& p \leftarrow a(X), \naf q(X/X), X \neq 0.
\end{align*}
is invariant under undefined arithmetics,
since substitutions that are not well-formed cannot yield applicable ground rules.
Hence, a vanilla grounder as considered above may skip the arithmetic evaluation
of ground terms obtained from $X/X$ without risking wrong answer sets.

\newpage
\section{EBNF Grammar and Lexical Table} \label{sec:gram}\label{sec:lex}

\begin{center}
\begin{framed}
\begin{verbatim}
<program>            ::= [<statements>] [<query>]

<statements>         ::= [<statements>] <statement>
<query>              ::= <classical_literal> QUERY_MARK

<statement>          ::= CONS [<body>] DOT
                       | <head> [CONS [<body>]] DOT
                       | WCONS [<body>] DOT 
                           SQUARE_OPEN <weight_at_level> SQUARE_CLOSE

<head>               ::= <disjunction> | <choice>
<body>               ::= [<body> COMMA]
                           (<naf_literal> | [NAF] <aggregate>)

<disjunction>        ::= [<disjunction> OR] <classical_literal>

<choice>             ::= [<term> <binop>] 
                           CURLY_OPEN [<choice_elements>]
                           CURLY_CLOSE [<binop> <term>]
<choice_elements>    ::= [<choice_elements> SEMICOLON] 
                           <choice_element>
<choice_element>     ::= <classical_literal> [COLON [<naf_literals>]]

<aggregate>          ::= [<term> <binop>] <aggregate function> 
                           CURLY_OPEN  [<aggregate_elements>]
                           CURLY_CLOSE [<binop> <term>]
<aggregate_elements> ::= [<aggregate_elements> SEMICOLON] 
                           <aggregate_element>
<aggregate_element>  ::= [<basic_terms>] [COLON [<naf_literals>]]
<aggregate_function> ::= AGGREGATE_COUNT
                       | AGGREGATE_MAX
                       | AGGREGATE_MIN 
                       | AGGREGATE_SUM

<weight_at_level>    ::= <term> [AT <term>] [COMMA <terms>]

<naf_literals>       ::= [<naf_literals> COMMA] <naf_literal>
<naf_literal>        ::= [NAF] <classical_literal> | <builtin_atom>

<classical_literal>  ::= [MINUS] ID [PAREN_OPEN [<terms>] PAREN_CLOSE]
<builtin_atom>       ::= <term> <binop> <term>

<binop>              ::= EQUAL
                       | UNEQUAL
                       | LESS
                       | GREATER
                       | LESS_OR_EQ
                       | GREATER_OR_EQ

<terms>              ::= [<terms> COMMA] <term>
<term>               ::= ID [PAREN_OPEN [<terms>] PAREN_CLOSE]
                       | NUMBER
                       | STRING
                       | VARIABLE
                       | ANONYMOUS_VARIABLE
                       | PAREN_OPEN <term> PAREN_CLOSE
                       | MINUS <term>
                       | <term> <arithop> term>

<basic_terms>        ::= [<basic_terms> COMMA] <basic_term>
<basic_term>         ::= <ground_term> |
                         <variable_term>
<ground_term>        ::= SYMBOLIC_CONSTANT |
                         STRING | [MINUS] NUMBER
<variable_term>      ::= VARIABLE |
                         ANONYMOUS_VARIABLE
<arithop>            ::= PLUS
                       | MINUS
                       | TIMES
                       | DIV
\end{verbatim}
\end{framed}
\end{center}


\newpage

\def\un{\char`\_}
\def\tb{\char`\\}
\def\te{\raisebox{-3pt}{\~{}}} 

\begin{small}
\begin{tt}
\begin{center}
\begin{longtable}{|l|l|l|}
    \hline
     {{\rm\small Token Name}} & {{\rm\small Mathematical Notation} }         & {{\rm\small Lexical Format} }    \\
     & {{\rm\small used within this document (exemplified)} }  & {{\rm(Flex Notation)}}\\
    \hline
        ID                         & $a, b, \mathit{anna}, \dots$                   & [a-z][A-Za-z0-9{\un}]* \\
        VARIABLE                   & $X, Y, \mathit{Name}, \dots$                   & [A-Z][A-Za-z0-9{\un}]* \\
        STRING                     & \rm ``http://bit.ly/cw6lDS'', ``Peter'', \dots & \tb "([\^{}\tb "]|\tb \tb \tb ")*\tb " \\
        NUMBER                     & $1, 0, 100000, \dots$                          & "0"|[1-9][0-9]* \\
        ANONYMOUS{\un}VARIABLE     & $\_$                                           & "{\un}" \\
        DOT                        & $.$                                            & "." \\
        COMMA                      & $,$                                            & "," \\
        QUERY{\un}MARK             & $?$                                            & "?" \\
        COLON                      & $:$                                            & ":" \\
        SEMICOLON                  & $;$                                            & ";" \\
        OR                         & $|$                                            & "|" \\
        NAF                        & $\naf$                                         & "not" \\
        CONS                       & $\leftarrow$                                   & ":-" \\
        WCONS                      & $\colonsim$                                    & ":\te" \\
        PLUS                       & $+$                                            & "+" \\
        MINUS                      & \rm $-$ or $\neg$                              & "-" \\
        TIMES                      & $*$                                            & "*" \\
        DIV                        & $/$                                            & "/" \\
        AT                         & $@$                                            & "@" \\
        PAREN{\un}OPEN             & $($                                            & "(" \\
        PAREN{\un}CLOSE            & $)$                                            & ")" \\
        SQUARE{\un}OPEN            & $[$                                            & "[" \\
        SQUARE{\un}CLOSE           & $]$                                            & "]" \\
        CURLY{\un}OPEN             & $\{$                                           & "\char`\{" \\
        CURLY{\un}CLOSE            & $\}$                                           & "\char`\}" \\
        EQUAL                      & $=$                                            & "=" \\
        UNEQUAL                    & $\not=$                                        & "<>"|"!=" \\
        LESS                       & $<$                                            & "<" \\
        GREATER                    & $>$                                            & ">" \\
        LESS{\un}OR{\un}EQ         & $\leq$                                         & "<=" \\
        GREATER{\un}OR{\un}EQ      & $\geq$                                         & ">=" \\
        AGGREGATE{\un}COUNT        & $\aggr{count}$                                 & "\#count" \\
        AGGREGATE{\un}MAX          & $\aggr{max}$                                   & "\#max" \\
        AGGREGATE{\un}MIN          & $\aggr{min}$                                   & "\#min" \\
        AGGREGATE{\un}SUM          & $\aggr{sum}$                                   & "\#sum" \\
        COMMENT                    & \rm \% this is a comment                       & "\%"([\^{}*\tb n][\^{}\tb n]*)?\tb n \\
        MULTI{\un}LINE{\un}COMMENT & \rm \%* this is a comment *\%                  & "\%*"([\^{}*]|\tb*[\^{}\%])*"*\%" \\
        BLANK                      &                                                & [ \tb t\tb n]+   \\

\hline
\end{longtable}
\end{center}
\end{tt}
\end{small}

Lexical values are given in Flex\footnote{\url{http://flex.sourceforge.net/}} syntax.
The {\tt COMMENT}, {\tt MULTI{\un}LINE{\un}COMMENT} and {\tt BLANK} tokens can be freely interspersed amidst other tokens
and have no syntactic   or semantic meaning.


\section{Conclusions}

In this document we have presented the \langname standard language that defines syntax and semantics of a standard language to which ASP solvers have to adhere in order to enter the ASP Competitions series, since 2013.
The standardization committee is still working on the evolution of the language in order to keep it aligned with the achievements of the ASP research community.
Among the features that are currently under consideration 
we mention here 
a semantics for recursive aggregates, for which several proposals are at the moment in place, e.g., \cite{pelo-2004,fabe-etal-2011-aij,alvi-etal-2011-jair,gelf-zhan-2014-tplp,alvi-etal-2015-tplp}, and 
a standard for intermediate \cite{gebs-etal-2016-iclp} and output \cite{brai-etal-2007-sea,outputformat} formats for ASP solvers.

\bibliographystyle{acmtrans}
\bibliography{bibtex}

\newcommand{\SortNoOp}[1]{}
\begin{thebibliography}{}

\bibitem[\protect\citeauthoryear{Abiteboul, Hull, and Vianu}{Abiteboul
  et~al\mbox{.}}{1995}]{abit-etal-95}
{\sc Abiteboul, S.}, {\sc Hull, R.}, {\sc and} {\sc Vianu, V.} 1995.
\newblock {\em {Foundations of Databases}}.
\newblock Addison-Wesley.

\bibitem[\protect\citeauthoryear{Alviano, Calimeri, Faber, Leone, and
  Perri}{Alviano et~al\mbox{.}}{2011}]{alvi-etal-2011-jair}
{\sc Alviano, M.}, {\sc Calimeri, F.}, {\sc Faber, W.}, {\sc Leone, N.}, {\sc
  and} {\sc Perri, S.} 2011.
\newblock Unfounded sets and well-founded semantics of answer set programs with
  aggregates.
\newblock {\em {Journal of Artificial Intelligence Research}\/}~{\em 42},
  487--527.

\bibitem[\protect\citeauthoryear{Alviano, Faber, and Gebser}{Alviano
  et~al\mbox{.}}{2015}]{alvi-etal-2015-tplp}
{\sc Alviano, M.}, {\sc Faber, W.}, {\sc and} {\sc Gebser, M.} 2015.
\newblock Rewriting recursive aggregates in answer set programming: back to
  monotonicity.
\newblock {\em {Theory and Practice of Logic Programming}\/}~{\em 15,\/}~4-5,
  559--573.

\bibitem[\protect\citeauthoryear{Brain, Faber, Maratea, Polleres, Schaub, and
  Schindlauer}{Brain et~al\mbox{.}}{2007}]{brai-etal-2007-sea}
{\sc Brain, M.}, {\sc Faber, W.}, {\sc Maratea, M.}, {\sc Polleres, A.}, {\sc
  Schaub, T.}, {\sc and} {\sc Schindlauer, R.} 2007.
\newblock What should an asp solver output? a multiple position paper.
\newblock In {\em Proc. of the First International SEA'07 Workshop}. CEUR
  Workshop Proceedings, vol. 281.

\bibitem[\protect\citeauthoryear{Brewka, Eiter, and Truszczy{\'n}ski}{Brewka
  et~al\mbox{.}}{2011}]{BrewkaET11}
{\sc Brewka, G.}, {\sc Eiter, T.}, {\sc and} {\sc Truszczy{\'n}ski, M.} 2011.
\newblock Answer set programming at a glance.
\newblock {\em Commun. ACM\/}~{\em 54,\/}~12, 92--103.

\bibitem[\protect\citeauthoryear{Calimeri, Cozza, Ianni, and Leone}{Calimeri
  et~al\mbox{.}}{2011}]{DBLP:journals/aicom/CalimeriCIL11}
{\sc Calimeri, F.}, {\sc Cozza, S.}, {\sc Ianni, G.}, {\sc and} {\sc Leone, N.}
  2011.
\newblock Finitely recursive programs: Decidability and bottom-up computation.
\newblock {\em {AI} Commun.\/}~{\em 24,\/}~4, 311--334.

\bibitem[\protect\citeauthoryear{Calimeri, Gebser, Maratea, and Ricca}{Calimeri
  et~al\mbox{.}}{2016}]{CalimeriGMR16}
{\sc Calimeri, F.}, {\sc Gebser, M.}, {\sc Maratea, M.}, {\sc and} {\sc Ricca,
  F.} 2016.
\newblock {Design and results of the Fifth Answer Set Programming Competition}.
\newblock {\em Artificial Intelligence\/}~{\em 231}, 151--181.

\bibitem[\protect\citeauthoryear{Calimeri, Ianni, and Ricca}{Calimeri
  et~al\mbox{.}}{2014}]{CalimeriIR14}
{\sc Calimeri, F.}, {\sc Ianni, G.}, {\sc and} {\sc Ricca, F.} 2014.
\newblock The third open answer set programming competition.
\newblock {\em {TPLP}\/}~{\em 14,\/}~1, 117--135.

\bibitem[\protect\citeauthoryear{Calimeri, Ianni, Ricca, and della Calabria
  Organizing~Committee}{Calimeri et~al\mbox{.}}{2011}]{languageformat}
{\sc Calimeri, F.}, {\sc Ianni, G.}, {\sc Ricca, F.}, {\sc and} {\sc della
  Calabria Organizing~Committee, T.~U.} 2011.
\newblock {Third ASP Competition, File and language formats}.
\newblock \url
  {http://www.mat.unical.it/aspcomp2011/files/LanguageSpecifications.pdf}.

\bibitem[\protect\citeauthoryear{Dell'Armi, Faber, Ielpa, Leone, and
  Pfeifer}{Dell'Armi et~al\mbox{.}}{2003}]{dell-etal-2003b}
{\sc Dell'Armi, T.}, {\sc Faber, W.}, {\sc Ielpa, G.}, {\sc Leone, N.}, {\sc
  and} {\sc Pfeifer, G.} 2003.
\newblock {Aggregate Functions in DLV}.
\newblock In {\em {Proceedings ASP03 - Answer Set Programming: Advances in
  Theory and Implementation}}, {M.~{de Vos}} {and} {A.~Provetti}, Eds. Messina,
  Italy, 274--288.
\newblock Online at \url{http://CEUR-WS.org/Vol-78/}.

\bibitem[\protect\citeauthoryear{Denecker, Pelov, and Bruynooghe}{Denecker
  et~al\mbox{.}}{2001}]{dene-etal-2001}
{\sc Denecker, M.}, {\sc Pelov, N.}, {\sc and} {\sc Bruynooghe, M.} 2001.
\newblock {Ultimate Well-Founded and Stable Model Semantics for Logic Programs
  with Aggregates}.
\newblock In {\em Proceedings of the 17th International Conference on Logic
  Programming}, {P.~Codognet}, Ed. Springer Verlag, 212--226.

\bibitem[\protect\citeauthoryear{Erdem, Gelfond, and Leone}{Erdem
  et~al\mbox{.}}{2016}]{ErdemGL16}
{\sc Erdem, E.}, {\sc Gelfond, M.}, {\sc and} {\sc Leone, N.} 2016.
\newblock Applications of answer set programming.
\newblock {\em {AI} Magazine\/}~{\em 37,\/}~3, 53--68.

\bibitem[\protect\citeauthoryear{Faber, Leone, and Pfeifer}{Faber
  et~al\mbox{.}}{2004}]{fabe-etal-2004-jelia}
{\sc Faber, W.}, {\sc Leone, N.}, {\sc and} {\sc Pfeifer, G.} 2004.
\newblock Recursive aggregates in disjunctive logic programs: Semantics and
  complexity.
\newblock In {\em {Proceedings of the 9th European Conference on Artificial
  Intelligence (JELIA 2004)}}, {J.~J. Alferes} {and} {J.~Leite}, Eds. {Lecture
  Notes in AI (LNAI)}, vol. 3229. Springer Verlag, 200--212.

\bibitem[\protect\citeauthoryear{Faber, Leone, and Pfeifer}{Faber
  et~al\mbox{.}}{2011}]{fabe-etal-2011-aij}
{\sc Faber, W.}, {\sc Leone, N.}, {\sc and} {\sc Pfeifer, G.} 2011.
\newblock Semantics and complexity of recursive aggregates in answer set
  programming.
\newblock {\em {Artificial Intelligence}\/}~{\em 175,\/}~1, 278--298.
\newblock Special Issue: John McCarthy's Legacy.

\bibitem[\protect\citeauthoryear{Ferraris}{Ferraris}{2005}]{ferr-2005-lpnmr}
{\sc Ferraris, P.} 2005.
\newblock {Answer Sets for Propositional Theories}.
\newblock In {\em {Logic Programming and Nonmonotonic Reasoning --- 8th
  International Conference, LPNMR'05, Diamante, Italy, September 2005,
  Proceedings}}, {C.~Baral}, {G.~Greco}, {N.~Leone}, {and} {G.~Terracina}, Eds.
  {Lecture Notes in Computer Science}, vol. 3662. Springer Verlag, 119--131.

\bibitem[\protect\citeauthoryear{Gebser, Harrison, Kaminski, Lifschitz, and
  Schaub}{Gebser et~al\mbox{.}}{2015}]{gehakalisc15a}
{\sc Gebser, M.}, {\sc Harrison, A.}, {\sc Kaminski, R.}, {\sc Lifschitz, V.},
  {\sc and} {\sc Schaub, T.} 2015.
\newblock Abstract {G}ringo.
\newblock {\em {Theory and Practice of Logic Programming}\/}~{\em 15,\/}~4-5,
  449--463.

\bibitem[\protect\citeauthoryear{Gebser, Kaminski, Kaufmann, Ostrowski, Schaub,
  and Wanko}{Gebser et~al\mbox{.}}{2016}]{gebs-etal-2016-iclp}
{\sc Gebser, M.}, {\sc Kaminski, R.}, {\sc Kaufmann, B.}, {\sc Ostrowski, M.},
  {\sc Schaub, T.}, {\sc and} {\sc Wanko, P.} 2016.
\newblock Theory solving made easy with clingo 5.
\newblock In {\em Technical Communications of the 32nd International Conference
  on Logic Programming (ICLP 2016 TCs)}, {M.~Carro}, {A.~King}, {N.~Saeedloei},
  {and} {M.~D. Vos}, Eds. {OASICS}, vol.~52. Schloss Dagstuhl - Leibniz-Zentrum
  fuer Informatik, 2:1--2:15.

\bibitem[\protect\citeauthoryear{Gebser, Leone, Maratea, Perri, Ricca, and
  Schaub}{Gebser et~al\mbox{.}}{2018}]{GebserLMPRS18}
{\sc Gebser, M.}, {\sc Leone, N.}, {\sc Maratea, M.}, {\sc Perri, S.}, {\sc
  Ricca, F.}, {\sc and} {\sc Schaub, T.} 2018.
\newblock Evaluation techniques and systems for answer set programming: a
  survey.
\newblock In {\em Proceedings of the Twenty-Seventh International Joint
  Conference on Artificial Intelligence ({IJCAI} 2018)}, {J.~Lang}, Ed.
  ijcai.org, 5450--5456.

\bibitem[\protect\citeauthoryear{Gebser, Maratea, and Ricca}{Gebser
  et~al\mbox{.}}{2017a}]{GebserMR17b}
{\sc Gebser, M.}, {\sc Maratea, M.}, {\sc and} {\sc Ricca, F.} 2017a.
\newblock The design of the seventh answer set programming competition.
\newblock In {\em Proceedings of the 14th International Conference on Logic
  Programming and Nonmonotonic Reasoning ({LPNMR} 2017)}, {M.~Balduccini} {and}
  {T.~Janhunen}, Eds. Lecture Notes in Computer Science, vol. 10377. Springer,
  3--9.

\bibitem[\protect\citeauthoryear{Gebser, Maratea, and Ricca}{Gebser
  et~al\mbox{.}}{2017b}]{GebserMR17}
{\sc Gebser, M.}, {\sc Maratea, M.}, {\sc and} {\sc Ricca, F.} 2017b.
\newblock The sixth answer set programming competition.
\newblock {\em Journal of Artificial Intelligence Research\/}~{\em 60}, 41--95.

\bibitem[\protect\citeauthoryear{Gebser, Obermeier, Schaub, Ratsch{-}Heitmann,
  and Runge}{Gebser et~al\mbox{.}}{2018}]{GebserOSR18}
{\sc Gebser, M.}, {\sc Obermeier, P.}, {\sc Schaub, T.}, {\sc
  Ratsch{-}Heitmann, M.}, {\sc and} {\sc Runge, M.} 2018.
\newblock Routing driverless transport vehicles in car assembly with answer set
  programming.
\newblock {\em {TPLP}\/}~{\em 18,\/}~3-4, 520--534.

\bibitem[\protect\citeauthoryear{Gelfond}{Gelfond}{2002}]{gelf-2002}
{\sc Gelfond, M.} 2002.
\newblock {Representing Knowledge in A-Prolog}.
\newblock In {\em {Computational Logic. Logic Programming and Beyond}}, {A.~C.
  Kakas} {and} {F.~Sadri}, Eds. LNCS, vol. 2408. Springer, 413--451.

\bibitem[\protect\citeauthoryear{Gelfond and Lifschitz}{Gelfond and
  Lifschitz}{1991}]{gelf-lifs-91}
{\sc Gelfond, M.} {\sc and} {\sc Lifschitz, V.} 1991.
\newblock {Classical Negation in Logic Programs and Disjunctive Databases}.
\newblock {\em {New Generation Computing}\/}~{\em 9}, 365--385.

\bibitem[\protect\citeauthoryear{Gelfond and Zhang}{Gelfond and
  Zhang}{2014}]{gelf-zhan-2014-tplp}
{\sc Gelfond, M.} {\sc and} {\sc Zhang, Y.} 2014.
\newblock Vicious circle principle and logic programs with aggregates.
\newblock {\em {Theory and Practice of Logic Programming}\/}~{\em 14,\/}~4-5,
  587--601.

\bibitem[\protect\citeauthoryear{Harrison and Lifschitz}{Harrison and
  Lifschitz}{2019}]{DBLP:journals/tplp/HarrisonL19}
{\sc Harrison, A.} {\sc and} {\sc Lifschitz, V.} 2019.
\newblock Relating two dialects of answer set programming.
\newblock {\em {TPLP}\/}~{\em 19,\/}~5-6, 1006--1020.

\bibitem[\protect\citeauthoryear{Kemp and Stuckey}{Kemp and
  Stuckey}{1991}]{kemp-stuc-91}
{\sc Kemp, D.~B.} {\sc and} {\sc Stuckey, P.~J.} 1991.
\newblock {Semantics of Logic Programs with Aggregates}.
\newblock In {\em {Proceedings of the International Symposium on Logic
  Programming (ISLP'91)}}, {V.~A. Saraswat} {and} {K.~Ueda}, Eds. {MIT Press},
  387--401.

\bibitem[\protect\citeauthoryear{Krennwallner}{Krennwallner}{2013}]{outputformat}
{\sc Krennwallner, T.} 2013.
\newblock {ASP Competition, output format}.
\newblock \url {https://www.mat.unical.it/aspcomp2013/files/aspoutput.txt}.

\bibitem[\protect\citeauthoryear{Leone and Ricca}{Leone and
  Ricca}{2015}]{leon-ricc-2015-rr}
{\sc Leone, N.} {\sc and} {\sc Ricca, F.} 2015.
\newblock Answer set programming: {A} tour from the basics to advanced
  development tools and industrial applications.
\newblock In {\em Web Logic Rules - 11th International Summer School on
  Reasoning Web, Tutorial Lectures}, {W.~Faber} {and} {A.~Paschke}, Eds.
  Lecture Notes in Computer Science, vol. 9203. Springer, 308--326.

\bibitem[\protect\citeauthoryear{Lierler, Maratea, and Ricca}{Lierler
  et~al\mbox{.}}{2016}]{LierlerMR16}
{\sc Lierler, Y.}, {\sc Maratea, M.}, {\sc and} {\sc Ricca, F.} 2016.
\newblock Systems, engineering environments, and competitions.
\newblock {\em {AI} Magazine\/}~{\em 37,\/}~3, 45--52.

\bibitem[\protect\citeauthoryear{Osorio and Jayaraman}{Osorio and
  Jayaraman}{1999}]{osor-jaya-99}
{\sc Osorio, M.} {\sc and} {\sc Jayaraman, B.} 1999.
\newblock {Aggregation and Negation-As-Failure}.
\newblock {\em {New Generation Computing}\/}~{\em 17,\/}~3, 255--284.

\bibitem[\protect\citeauthoryear{Pelov}{Pelov}{2004}]{pelo-2004}
{\sc Pelov, N.} 2004.
\newblock {Semantics of Logic Programs with Aggregates}.
\newblock Ph.D. thesis, Katholieke Universiteit Leuven, Leuven, Belgium.

\bibitem[\protect\citeauthoryear{Pelov, Denecker, and Bruynooghe}{Pelov
  et~al\mbox{.}}{2004}]{pelo-etal-2004a}
{\sc Pelov, N.}, {\sc Denecker, M.}, {\sc and} {\sc Bruynooghe, M.} 2004.
\newblock Partial stable models for logic programs with aggregates.
\newblock In {\em {Proceedings of the 7th International Conference on Logic
  Programming and Non-Monotonic Reasoning (LPNMR-7)}}. {Lecture Notes in AI
  (LNAI)}, vol. 2923. Springer, 207--219.

\bibitem[\protect\citeauthoryear{Pelov, Denecker, and Bruynooghe}{Pelov
  et~al\mbox{.}}{2007}]{pelo-etal-2007-tplp}
{\sc Pelov, N.}, {\sc Denecker, M.}, {\sc and} {\sc Bruynooghe, M.} 2007.
\newblock {Well-founded and Stable Semantics of Logic Programs with
  Aggregates}.
\newblock {\em {Theory and Practice of Logic Programming}\/}~{\em 7,\/}~3,
  301--353.

\bibitem[\protect\citeauthoryear{Pelov and Truszczy{\'n}ski}{Pelov and
  Truszczy{\'n}ski}{2004}]{pelo-trus-2004}
{\sc Pelov, N.} {\sc and} {\sc Truszczy{\'n}ski, M.} 2004.
\newblock Semantics of disjunctive programs with monotone aggregates - an
  operator-based approach.
\newblock In {\em Proceedings of the 10th International Workshop on
  Non-monotonic Reasoning (NMR 2004), Whistler, BC, Canada}. 327--334.

\bibitem[\protect\citeauthoryear{Ross and Sagiv}{Ross and
  Sagiv}{1997}]{ross-sagi-1997}
{\sc Ross, K.~A.} {\sc and} {\sc Sagiv, Y.} 1997.
\newblock {Monotonic Aggregation in Deductive Databases}.
\newblock {\em {Journal of Computer and System Sciences}\/}~{\em 54,\/}~1
  (Feb.), 79--97.

\bibitem[\protect\citeauthoryear{Simons, Niemel{\"a}, and Soininen}{Simons
  et~al\mbox{.}}{2002}]{simo-etal-2002}
{\sc Simons, P.}, {\sc Niemel{\"a}, I.}, {\sc and} {\sc Soininen, T.} 2002.
\newblock {Extending and Implementing the Stable Model Semantics}.
\newblock {\em {Artificial Intelligence}\/}~{\em 138}, 181--234.

\bibitem[\protect\citeauthoryear{{Van Gelder}}{{Van Gelder}}{1992}]{vang-92}
{\sc {Van Gelder}, A.} 1992.
\newblock {The Well-Founded Semantics of Aggregation}.
\newblock In {\em {Proceedings of the Eleventh Symposium on Principles of
  Database Systems (PODS'92)}}. {ACM Press}, 127--138.

\end{thebibliography}

\label{lastpage}
\end{document}